\documentclass[10pt,twocolumn,letterpaper]{article}

\usepackage[pagenumbers]{iccv}              %
\usepackage{bm}
\usepackage{bbm}
\usepackage{makecell}

\usepackage{float}

\renewcommand{\paragraph}[1]{\noindent\textbf{#1}}
\usepackage{multirow}
\usepackage[utf8]{inputenc}
\usepackage{pifont}
\DeclareUnicodeCharacter{2713}{\checkmark}

\definecolor{iccvblue}{rgb}{0.21,0.49,0.74}
\usepackage[pagebackref,breaklinks,colorlinks,citecolor=iccvblue]{hyperref}

\def\paperID{2744} %
\def\confName{ICCV}
\def\confYear{2025}

\newcommand{\yuri}[1]{\textcolor{green}{[Yuri: #1]}}

\newcommand\blfootnote[1]{%
  \begingroup
  \renewcommand\thefootnote{}\footnote{#1}%
  \addtocounter{footnote}{-1}%
  \endgroup
}

\newcommand{\method}{SpatialTrackerV2\xspace}

\newcommand{\ego}{\mathcal{T}_{\textrm{ego}}}
\newcommand{\object}{\mathcal{T}_{\textrm{object}}}

\newcommand{\cam}{\ego}

\newcommand{\video}{(\mathcal{I}^t)_{t=1}^T}

\newcommand{\poset}{\mathbf{P}_{\textrm{tok}}}
\newcommand{\scalet}{\mathbf{S}_{\textrm{tok}}}

\newcommand{\warp}{{\mathcal{W}}}

\newcommand{\invdepth}{\mathcal{D}_{\textrm{norm}}}
\newcommand{\campose}{\mathcal{P}^t}

\newcommand{\depthcorr}{Corr_{\textrm{3D}}}

\title{\method: 3D Point Tracking Made Easy
}

\author{
    Yuxi Xiao\textsuperscript{1}$^{*}$ \quad
    Jianyuan Wang\textsuperscript{2} \quad
    Nan Xue\textsuperscript{3} \quad
    Nikita Karaev\textsuperscript{2,4} \quad
    Yuri Makarov\textsuperscript{4} \quad
    Bingyi Kang\textsuperscript{5} \\[2pt]
    Xing Zhu\textsuperscript{3} \quad
    Hujun Bao\textsuperscript{1} \quad
    Yujun Shen\textsuperscript{3} \quad
    Xiaowei Zhou\textsuperscript{1}$^{\dagger}$ \quad
    \\[5pt]
    $^1$Zhejiang University \qquad
    $^2$Oxford \qquad
    $^3$Ant Group \qquad
    $^4$Pixelwise AI \qquad
    $^5$Bytedance Seed
    \\[8pt]
}

\begin{document}

\twocolumn[{
\renewcommand\twocolumn[1][]{#1}
\maketitle
\begin{center}
    \vspace{-28pt}
    \includegraphics[width=0.97\linewidth]{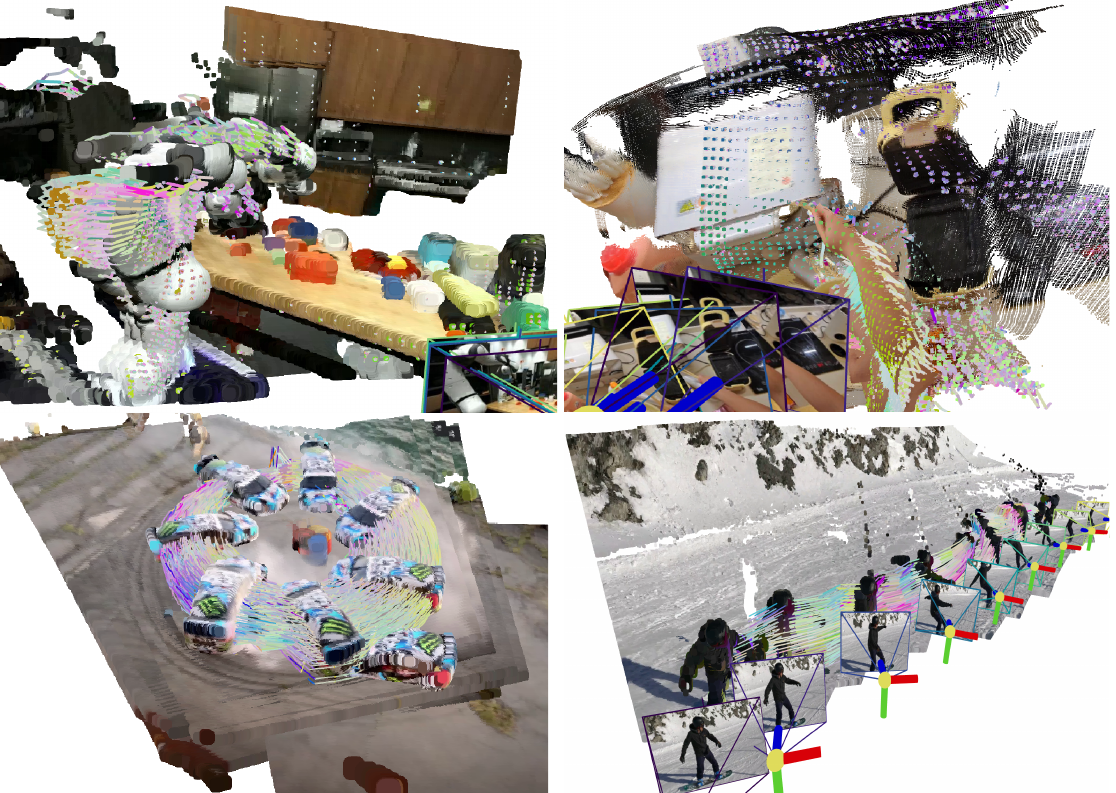}
    \captionsetup{type=figure}
    \vspace{-5pt}
    \caption{\textbf{SpatialTrackerV2} produces consistent 3D scene geometry, camera poses, and 3D point trajectories all at once from monocular videos of arbitrary scenarios, e.g., robotic manipulation, first-person egocentric views, and dynamic sports (drifting and skating) shown in this figure. Try our online demo at { \url{https://huggingface.co/spaces/Yuxihenry/SpatialTrackerV2}.
        }
    }
    \label{fig:teaser}
\end{center}
}]

\blfootnote{\hspace{-2em}$^{*}$ Partially completed during Ant internship. $^{\dagger}$ Corresponding author.}

\vspace{-1.5em}
\begin{abstract}
We present \method, a feed-forward 3D point tracking method for monocular videos. 
Going beyond modular pipelines built on off-the-shelf components for 3D tracking, our approach unifies the intrinsic connections between point tracking, monocular depth, and camera pose estimation into a high-performing and feedforward 3D point tracker.
It decomposes world-space 3D motion into scene geometry, camera ego-motion, and pixel-wise object motion, with a fully differentiable and end-to-end architecture, allowing scalable training across a wide range of datasets, including synthetic sequences, posed RGB-D videos, and unlabeled in-the-wild footage. 
By learning geometry and motion jointly from such heterogeneous data, \method outperforms existing 3D tracking methods by {\bf 30\%}, and matches the accuracy of leading dynamic 3D reconstruction approaches while running 50× faster.
\end{abstract}
    
\section{Introduction}
\label{sec:intro}
3D point tracking aims to recover long-term 3D trajectories of arbitrary points from monocular videos. As a universal dynamic scene representation, it has recently shown strong potential in diverse applications, including robotics~\cite{Robot_GS, Kalib, niu2025pre}, video generation~\cite{wang2024objctrl, Diff_as_shader, track4Gen, ouyang2024codef}, and 3D/4D reconstruction~\cite{kasten2024fast, lei2024mosca, wang2024shape, PreF3R}. Compared to parametric motion models (e.g., SMPL~\cite{SMPL}, MANO~\cite{MANO}, skeletons, or 3D bounding boxes), it offers greater flexibility and generalization over various real-world scenes, as shown in Fig.~\ref{fig:teaser}.

Existing solutions~\cite{wang2023tracking,xiao2024spatialtracker,GS_Dit,ngo2024delta, badki2025l4p} of 3D point tracking extensively explored the well-developed low/mid-level vision models, such as optical flow~\cite{teed2020raft, liu2010sift} and monocular depth estimation~\cite{yang2024depth, depthcrafter}, and took benefits from 2D point tracking models~\cite{karaev2024cotracker, doersch2023tapir, harley2022particle}. Among them, optimization-based methods~\cite{Particlesfm,wang2023tracking,li2025megasam} distill the optical flow, monocular depth models and camera motions for each given monocular video with promising results obtained, while being computationally expensive due to their per-scene optimization designs. 
SpatialTracker~\cite{xiao2024spatialtracker} moved forward to efficient 3D point tracking with a feed-forward model, and the more recent works~\cite{ngo2024delta, badki2025l4p, GS_Dit} explored different architecture designs and rendering constraints to achieve higher-quality 3D tracking. Nevertherless, the feed-forward solutions are limited to training data scalability issues due to the need for ground-truth 3D tracks as supervision, which downgrades the tracking quality in real-world casual captures. Moreover, overlooking the inherent interplay between camera motion, object motion, and scene geometry results in error entanglement and accumulation across modules.

These limitations motivate our core insights:
(1) The reliance on ground-truth 3D trajectories constrains the scalability of existing feed-forward models, highlighting the need for designs that can generalize across diverse and weakly-supervised data sources. (2) The absence of joint reasoning over scene geometry, camera motion, and object motion leads to compounded errors and degraded performance, underscoring the importance of disentangling and explicitly modeling these motion components. To address these challenges, we decompose 3D point tracking into three distinct components: video depth, ego (camera) motion, and object motion, and integrate them within a fully differentiable pipeline that supports scalable joint training across heterogeneous data.

In our SpatialTrackerV2, a front-end and back-end architecture is proposed. The front end is a video depth estimator and camera pose initializer, adapted from typical monocular depth prediction frameworks~\cite{yang2025depth} with attention-based temporal information encoding~\cite{wang2025vggt}.
The predicted video depths and camera poses are then fused through a scale-shift estimation module, which ensures consistency between the depth and motion predictions. The back end consists of a proposed Joint Motion Optimization Module, which takes the video depth and coarse camera trajectories as input and iteratively estimates 2D and 3D trajectories, along with trajectory-wise dynamics and visibility scores. This enables an efficient bundle adjustment process for optimizing camera poses in the loop. At its core lies a novel SyncFormer, which separately models the 2D and 3D correlations in two branches, connected by multiple cross-attention layers. This design mitigates mutual interference between 2D and 3D embeddings and allows the model to update representations in two distinct spaces, namely the image (UV) space and the camera coordinate space. Furthermore, benefiting from this dual-branch design, bundle adjustment can be effectively applied to jointly optimize camera poses as well as the 2D and 3D trajectories.

This unified and differentiable pipeline makes large-scale training on diverse datasets possible. For RGB-D datasets with camera poses, we jointly train 3D tracking using consistency constraints from ground-truth depth and camera poses for static points, while dynamic points seamlessly contribute to the optimization. For video datasets that provide only camera pose annotations and lack depth information, we leverage consistency among camera poses, and 2D and 3D point tracking to drive the model’s optimization. Relying on this framework, we successfully scale up training of the entire pipeline across 17 datasets.

Evaluations on the TAPVid-3D benchmark~\cite{tap3d} show that our method sets a new state-of-the-art in 3D point tracking, achieving 21.2 AJ and 31.0 APD$_{3D}$, surpassing DELTA~\cite{ngo2024delta} with relative improvements of 61.8\% and 50.5\%, respectively.  
Additionally, extensive experiments on dynamic reconstruction show our superior results on consistent video depth and camera poses estimation. Specifically, SpatialTrackerV2 beats the best dynamic reconstruction method, MegaSAM~\cite{li2025megasam}, on most video depth datasets and achieves comparable results on various camera pose benchmarks, while its inference speed is 50 $\times$ faster.

\section{Related work}
\label{sec:formatting}

This section covers relevant literature on 3D tracking, depth, and camera pose estimation. 

\subsection{Point tracking}

PIPs~\cite{harley2022particle} revisited the 2D point tracking task first introduced in~\cite{sand2008particle} and proposed a deep learning approach to solve it. TAP-Vid~\cite{doersch2022tap} redefined the problem and introduced both a benchmark and a simple architecture, TAP-Net. Subsequently, the performance was improved in TAPIR~\cite{doersch2023tapir} by combining the global matching capabilities of TAP-Net with the local refinement offered by PIPs. CoTracker~\cite{karaev2024cotracker} pioneered tracking through occlusions using a transformer architecture combined with joint attention, followed by TAPTR~\cite{li2024taptr} and LocoTrack~\cite{cho2024local}, which improved efficiency and introduced 4D correlation volumes.
Recently, BootsTAPIR~\cite{doersch2024bootstap} and CoTracker3~\cite{karaev2024cotracker3} explored the use of unlabeled data to achieve better performance.

While 2D point tracking has been extensively studied, 3D point tracking remains a relatively new field. The first method to demonstrate 3D point tracking capabilities was the test-time optimization-based OmniMotion~\cite{wang2023tracking}.
Later, SpatialTracker~\cite{xiao2024spatialtracker} introduced the first feed-forward 3D point tracker by combining a 2D point tracker~\cite{karaev2024cotracker} with depth priors from a monocular depth estimator~\cite{bhat2023zoedepth}. SceneTracker~\cite{wang2024scenetracker} proposed a new architecture for 3D tracking with depth priors, while DELTA~\cite{ngo2024delta} improved efficiency and achieved dense 3D tracking. Recently, TAPIP3D~\cite{zhang2025tapip3d} improved 3D tracking robustness by lifting image features into the world coordinate space and performing tracking there. All of these 3D tracking models were trained on small synthetic datasets.

Unlike these methods, we present a scalable 3D tracking framework trained on a collection of both real and synthetic datasets, while also explicitly modeling camera motion to improve performance on egocentric videos.

\begin{figure*}
    \centering
    \includegraphics[width=\linewidth]{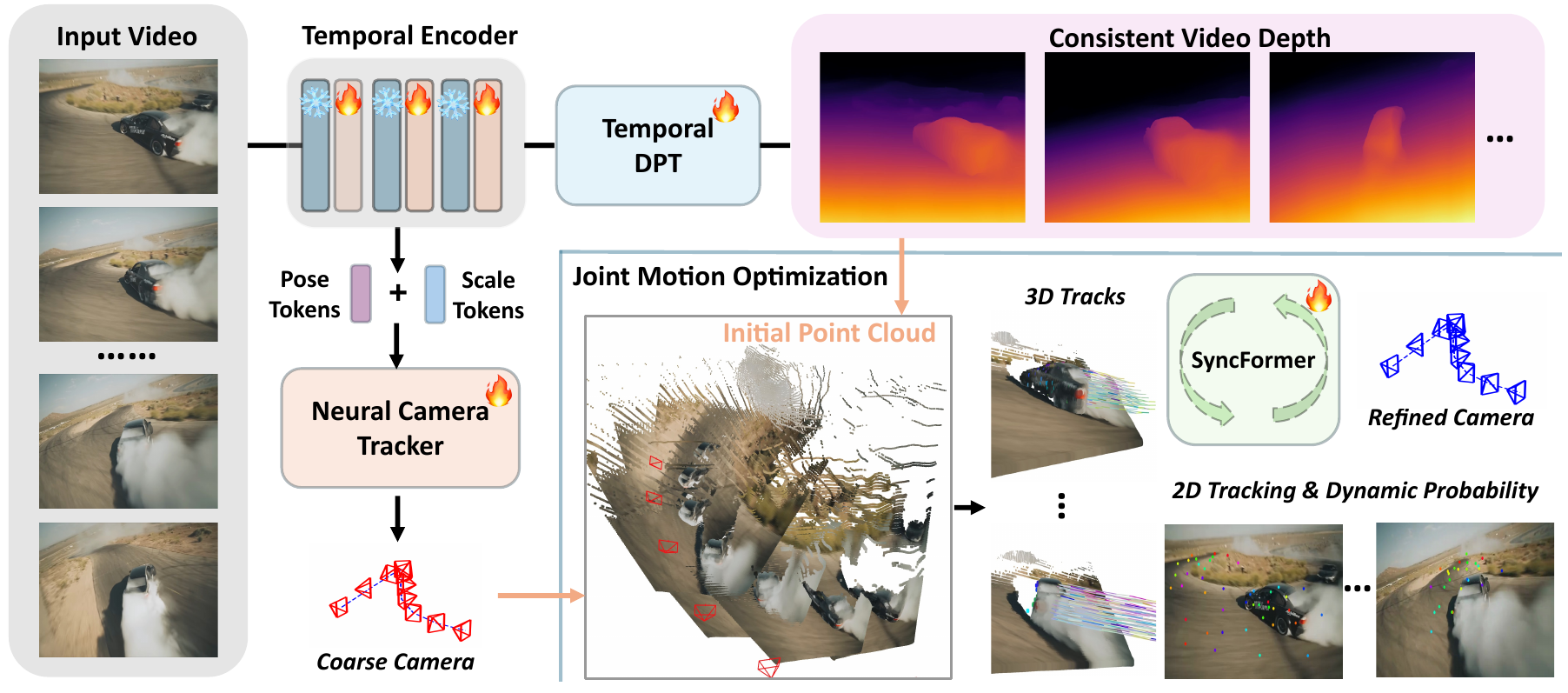}
    \caption{\textbf{Pipeline Overview.} Our method adopts a front-end and back-end architecture. The front-end estimates scale-aligned depth and camera poses from the input video, which are used to construct initial static 3D tracks. The back-end then iteratively refines both tracks and poses via joint motion optimization.}
    \label{fig:pipeline}
\end{figure*}

\subsection{Depth estimation}
Early methods such as Eigen et al.~\cite{Eigen} introduced single-view depth estimation using CNNs, but were limited by dataset scale and poor generalization~\cite{DEEPORD,Pingtan}. MiDaS~\cite{Midas} improved this by mixing datasets for broader coverage, and ZoeDepth~\cite{bhat2023zoedepth} adapted it for metric depth, though ambiguity from missing camera intrinsics remained. Later, Metric3D~\cite{metric3d} and UniDepth~\cite{UniDepth} addressed this by jointly estimating intrinsics and normalized depth.

With large-scale pretraining (e.g., diffusion~\cite{dhariwal2021diffusion, li2024drip}, DINO~\cite{caron2021dino}), recent models like Marigold~\cite{marigold} and DepthAnything~\cite{yang2024depth,yang2025depth} have significantly advanced zero-shot depth. Extensions to video~\cite{depthcrafter,videoDepth} have also emerged.

In this work, we build on DepthAnythingV2 and extend it to video, aiming for not just consistent depth, but a unified framework that also predicts camera poses and tracks, with proper scale alignment—posing new challenges beyond static depth estimation.

\subsection{Camera estimation}
Traditional camera pose estimation methods~\cite{hartley_multiple_2004, ozyecsil2017survey} rely on image-to-image point correspondences using keypoint detectors (e.g., SIFT~\cite{lowe_object_1999,lowe_distinctive_2004-1}, SURF~\cite{bay_speeded-up_2008}) and matching techniques such as nearest neighbors, followed by geometric algorithms like the five-point and eight-point methods~\cite{hartley_multiple_2004,hartley1997defense,li2006five,nister2004efficient, xiao2023level}. Bundle Adjustment~\cite{triggs2000bundle} is also commonly employed to further enhance accuracy.
Recently, direct regression approaches using neural networks~\cite{kehl2017ssd, xiang2017posecnn, ma2022virtual, zhang2022relpose, lin2023relposepp} have emerged, aiming to overcome limitations in sparse-view scenarios or when correspondences are unreliable. Diffusion models, such as PoseDiffusion~\cite{wang2023posediffusion} and RayDiffusion~\cite{zhang2024cameras}, have also been explored, offering strong accuracy but suffering from high inference costs. In contrast, the camera head of VGGSfM~\cite{wang2024vggsfm} or VGGT~\cite{wang2025vggt} adopts an iterative refinement paradigm similar to RAFT~\cite{teed2020raft}, striking a balance between accuracy and inference cost, and enabling estimation of both extrinsic and intrinsic parameters.
\if
\subsection{3D perception models}
\yuri{we have to think about proper name for this section}
dust3r, mast3r, CUT3R, MEGA-SAM

\yuri{Shape of Motion? (test time optimization but maybe relevant)}
\fi

\section{Method}

Given a video with $T$ frames $\video$ and $N$ query points $\mathcal{Q}_{i} = (x_i, y_i)\in \mathbb{R}^{2}, i = 1,\dots,N$, the goal of 3D tracking is to recover pixel-wise 3D trajectories $\mathcal{T} = (\mathcal{T}_i^t)_{i=1,\dots,N}^{t=1,\dots,T}$ for each query. In order to account for static and dynamic parts of the scene, we decompose $\mathcal{T}$ into the ego camera motion $\ego$ and object motion $\object$ as shown in \cref{fig:pipeline}.

\subsection{Ego Motion Component}
Ego motion, represented by camera trajectories $\cam$, is a major contributor to the 3D flow in camera coordinates. To compute the 3D tracks induced by ego motion, we need to estimate scale-aligned camera trajectories and video depth.

\paragraph{Video Depth.}
Monocular depth estimation models, such as DepthAnything~\cite{yang2024depth}, typically follow an encoder-decoder design with a vision encoder like DINO~\cite{caron2021dino} and a DPT-style decoder. Following VGGT~\cite{wang2025vggt}, we extend the monocular encoder into a temporal encoder using an alternating-attention mechanism. This mechanism alternates between intra-frame self-attention and inter-frame attention over flattened video tokens, effectively balancing performance and efficiency. Furthermore, two learnable tokens, $\poset$ and $\scalet$, are incorporated into the alternating-attention layers to capture high-level semantics for pose and scale regression.

\paragraph{Camera Tracker.}
\label{sec:cam}
Similar to ~\cite{wang2025vggt,wang2024vggsfm}, we adopt a differentiable pose head, $\mathcal{H}$, to decode pose, scale and shift directly:
\begin{equation}
    \campose, a, b = \mathcal{H}(\poset, \scalet),
\end{equation} where $\campose \in \mathbb{R}^{T\times8}$ is the camera encoding parameterized by a quaternion, a translation vector, and the normalized focal length concatenated together; The parameters $a$ and $b$ represent the scale and shift used to align the depth with the estimated camera poses. After that, we can easily obtain the 3D trajectories induced by ego-motion $\ego$:
\begin{equation}
    \ego = \warp(\mathcal{P}, a*\invdepth+b), %
\end{equation}
where $\invdepth$ is the raw results after DPT head with activation functions and $\warp$ is the camera transformation. 

\begin{figure}
\centering
\includegraphics[width=\linewidth]{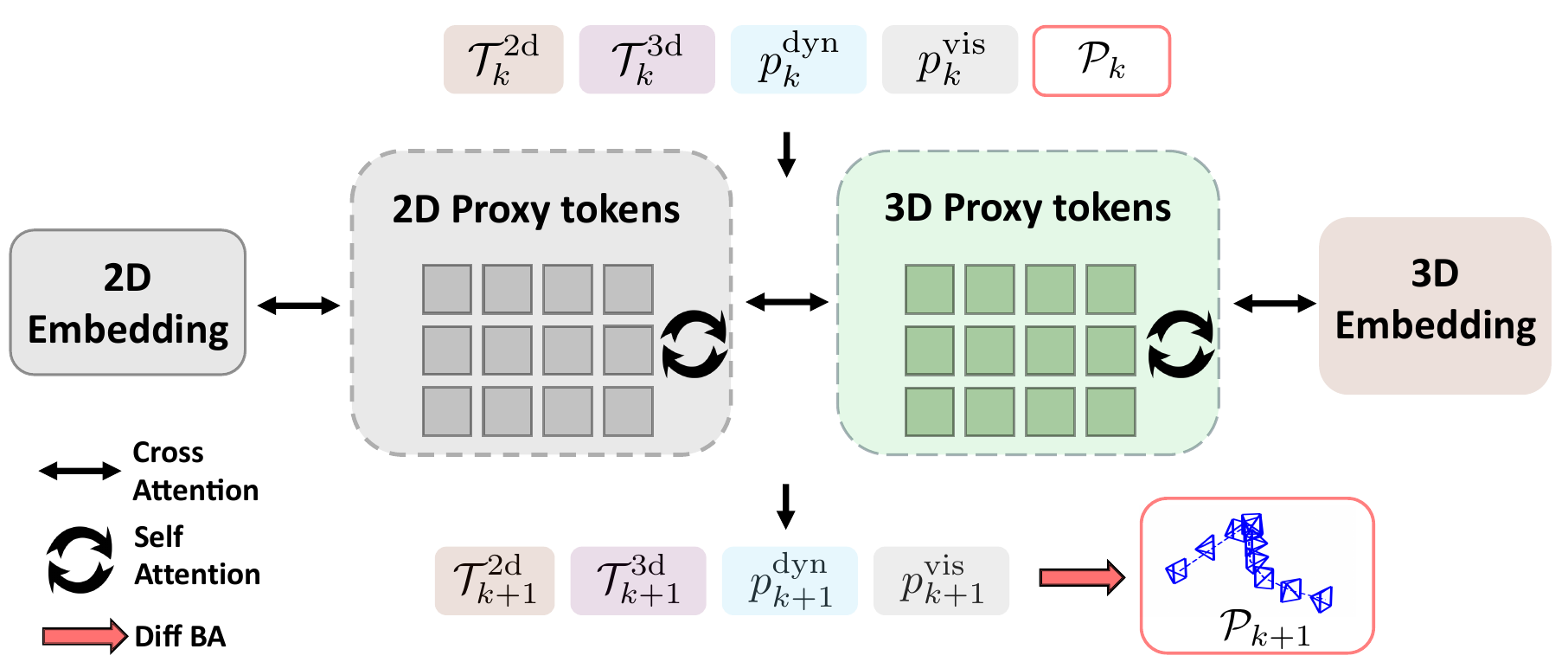}
\caption{
\textbf{SyncFormer.} The model takes previous estimates and their corresponding embeddings as input, and updates them iteratively. The 2D and 3D embeddings are processed in separate branches that interact via cross-attention.
}%
\label{fig:switcher}
\end{figure}

\subsection{Joint Motion Optimization.}
After ego-motion initialization, we jointly estimate 2D trajectories $\mathcal{T}^{\text{2d}} \in \mathbb{R}^{T \times N \times 2}$ in UV space and their corresponding 3D trajectories $\mathcal{T}^{\text{3d}} \in \mathbb{R}^{T \times N \times 3}$ in the camera coordinate system using an iterative transformer module, referred to as SyncFormer. In parallel, the SyncFormer also dynamically estimates the visibility probability $p^{\text{vis}}$ and dynamic probability $p^{\text{dyn}}$ for each trajectory, enabling an efficient bundle adjustment process to refine the camera poses $\mathcal{P}$. 

\paragraph{\textbf{SyncFormer.}} As illustrated in Fig.~\ref{fig:switcher}, in every iteration, SyncFormer takes 2D embeddings, 3D embeddings, and camera poses as input, and updates the 2D trajectories $\mathcal{T}^{\text{2d}}$, 3D trajectories $\mathcal{T}^{\text{3d}}$, dynamic probabilities $p^{\textrm{dyn}}$, and visibility scores $p^{\textrm{vis}}$:
\begin{equation}
\small 
    \mathcal{T}_{k+1}^{\text{2d}}, \mathcal{T}_{k+1}^{\text{3d}}, p^{\textrm{dyn}}_{k+1}, p^{\textrm{vis}}_{k+1} = f_{\rm sync}(\mathcal{T}_{k}^{\text{2d}}, 
    \mathcal{T}_{k}^{\text{3d}}, 
    p^{\textrm{dyn}}_{k}, p^{\textrm{vis}}_{k}, \mathcal{P}_{k}),
\end{equation}
where $f_{\rm sync}$ denotes the transformer-based update function. To better capture the distinct characteristics of 2D and 3D motion, the 2D and 3D trajectory updaters are modeled using separate attention layers. To reduce computational cost, correlation embeddings are first encoded into a compact set of 2D and 3D proxy tokens using cross-attention. Then, information is exchanged between the 2D and 3D branches via a cross-attention layer between the respective proxy tokens. This design decouples the mutual influence between 2D and 3D tracking, which are updated in two different domains: the UV space for 2D trajectories and the camera coordinate space for 3D trajectories. 

\paragraph{2D and 3D Embeddings.} As the input for SyncFormer, the 2D and 3D embeddings encode the status of current estimations while recording the neighbourhood information for updating. For 2D embeddings, we keep the same to Cotracker3~\cite{karaev2024cotracker}. The 3D embeddings $\mathbf{E}^{\text{3d}} = (\depthcorr, \textbf{e}^{\text{time}}, \textbf{e}^{\text{Gpos}}, p^{\text{dyn}}, p^{\text{vis}})$ contains the 3D correlation features $\depthcorr$, the time embeddings $\textbf{e}^{\text{time}}$, global position embeddings $\textbf{e}^{\text{Gpos}}$ and dynamic-visibility scores. The 3D correlation $\depthcorr$ is the main features for the 3D embedding. Different to 2D correlations, we expect 3D tracking branch updates the 3D position in the camera coordinate space. Therefore, we calculate the 3D correlations on the normalized point maps from front-end instead of depth map. Similar to 2D correlations, we construct multi-resolution point maps and compute the relative translations between each point and its neighbors within radius of 3. We then apply harmonic positional encoding to project these relative translations into high-dimensional feature representations, which are combined with semantic features to compute the final 3D correlations:
\begin{equation}
    \depthcorr = [K(\frac{\mathbf{x}}{ks}+\delta,\frac{\mathbf{y}}{ks}+\delta): \delta\in\mathbb{Z}, \|\delta\|_{\infty} \leq \Delta],
\end{equation} with $\Delta$ set to 3 and at multiple scales $s=1,2,3,4$, $K$ denotes the operator for encoding relative translation features. Besides, in order to encode the camera pose to assist the tracking estimations, we propose $\textbf{e}^{\text{Gpos}}$ which transforms the query points $\mathcal{Q}_{i}$ into each frames with the current camera poses, denoted as anchor points $\mathcal{Q}_{i}^{\text{anc}}$. $\textbf{e}^{\text{Gpos}}$ is calculated from the relative translation between the current position and $\textbf{e}^{\text{Gpos}}$, and they are then projected into high dimension with same tricks as above.

\begin{figure}
\centering
\includegraphics[width=\linewidth]{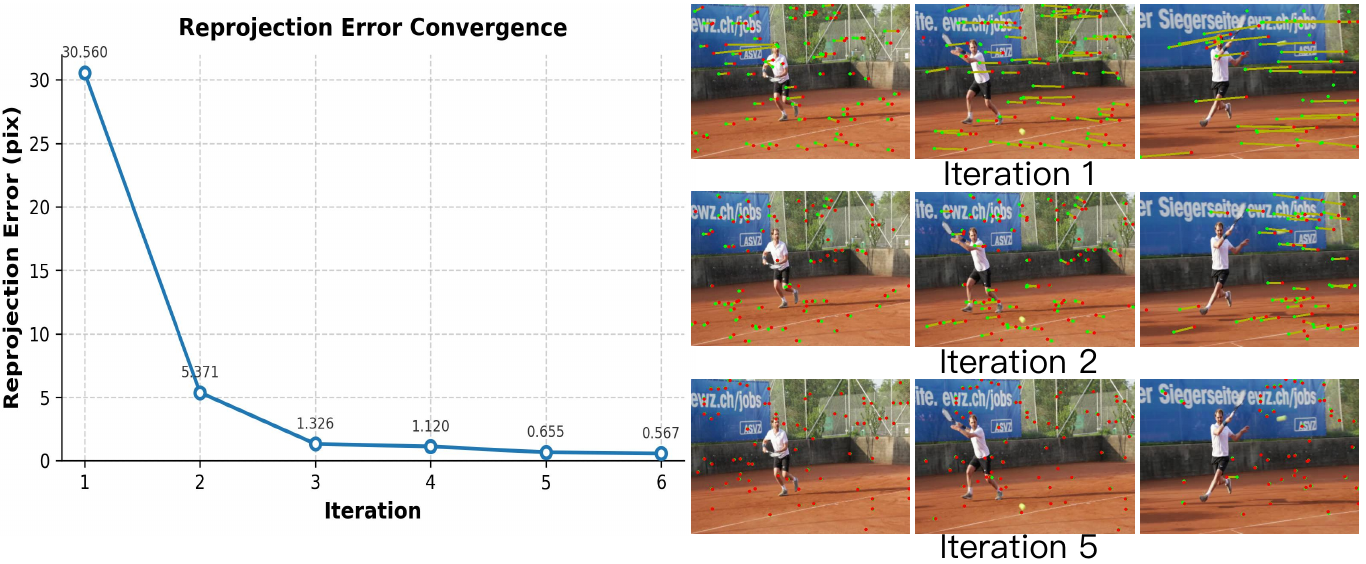}
\caption{
\textbf{Iterative Process of SyncFormer.} The left subfigure shows the convergence curve of reprojection error, illustrating rapid reprojection error reduction. The right subfigures visualizes the progressive alignment between 2D tracking results (in green) and projections of 3D (world) tracking transformed by camera poses (in red).
}
\label{fig:syncformer_iter}
\end{figure}
\paragraph{Camera Motion Optimization.}
After each iteration, the updates of $\mathcal{T}_{k+1}^{\text{2d}}, \mathcal{T}_{k+1}^{\text{3d}}, p^{\textrm{dyn}}_{k+1}, p^{\textrm{vis}}_{k+1}$ naturally form self-consistency constraints with the camera poses. These intrinsic constraints can be seamlessly incorporated into a bundle adjustment formulation. Specifically, we apply a weighted Procrustes analysis to register $\mathcal{T}_{k+1}^{\text{3d}}$ into the world coordinate frame, where the alignment weights are given by the dynamic scores. This coarse alignment process is differentiable and provides effective supervision for learning the dynamic probability. Then we fuse the aligned $\mathcal{T}_{k+1}^{\text{3d}}$ to world points $\mathbf{P}_{k+1}^{\text{world}} \in \mathbb{R}^{N\times 3}$, where the dynamic points are filtered with the estimated dynamic scores. After this, we apply a direct bundle adjustment to optimize the camera poses with paired $\mathbf{P}_{k+1}^{\text{world}}, \mathcal{T}_{k+1}^{\text{2d}}$ and $p^{\textrm{vis}}_{k+1}$. In the next iteration, the updated camera poses $\mathcal{P}_{k+1}$ influence the subsequent loop through global position embedding. It is worth noting that the 2D and 3D trajectories are not updated via bundle adjustment; instead, the entire system is primarily driven by SyncFormer. As shown in Fig.~\ref{fig:syncformer_iter}, the reprojection errors decrease rapidly over iterations, while the 2D trajectories gradually become consistent with the 3D projections transformed by camera motion.

\subsection{Training}
\label{sec:point}
\paragraph{Training Datasets.} Our model is trained on a large collection of 17 datasets, each providing different forms of supervision. The training data can be broadly categorized into three types: (1) Posed RGB-D with tracking annotations, (2) Posed RGB-D, (3) Pose-only or unlabeled data. For (1) containing Kubric~\cite{greff2022kubric}, PointOdyssey~\cite{zheng2023pointodyssey} and Dynamic Replica~\cite{karaev2023dynamicstereo}, we add the full supervisions of camera poses, video depth, dynamic segmentation and 2D, 3D tracking. Within category (2), including VKITTI~\cite{cabon2020vkitti2}, TartanAir~\cite{wang2020tartanair}, Spring~\cite{Mehl2023_Spring}, DL3DV~\cite{ling2024dl3dv}, BEDLAM~\cite{BELAM}, MVS-Synth~\cite{MVS_syn}, CO3Dv2~\cite{Co3DV2}, COP3D~\cite{COP3D}, WildRGBD~\cite{wildrgbd}, ScanNet++~\cite{dai2017scannet} and OmniObject3D~\cite{Omniobject}, we leverage depth supervision to improve the video depth estimator. In addition, joint training losses are applied to ensure that the 3D tracking remains consistent with the estimated depth. As for category (3), it includes HOI4D~\cite{HOI4D}, Ego4D~\cite{grauman2022ego4d}, and Stereo4D~\cite{jin2024stereo4d}. For this type of data, we apply camera loss and joint training losses here, and follow~\cite{videoDepth} by using a monocular depth model~\cite{wang2024moge} as a teacher to preserve relative depth accuracy. It is worth noting that Stereo4D~\cite{jin2024stereo4d} provides only very sparse depth on valid 3D tracks. Therefore, we choose not to use its annotations and instead treat it as pose-only data, given that it is sourced from Internet videos with rich scene diversity. 

\paragraph{Implementation details.} Our training recipe consists of several stages.
\textbf{Stage 1.} We first train the front end model to jointly estimate video depth and camera poses on category (1) and (2) datasets which sums up to 14 datasets. Training is conducted with mixed BF16 precision, while the DPT and camera tracker modules use full precision to ensure stable optimization. We use the AdamW optimizer with a learning rate of $5\times 10^{-5}$ and apply gradient clipping with a threshold of 0.1. Stage 1 training is performed for $200$k iterations on $64$ H20 GPUs. We shuffle the video length from 1-24 during the training.   
\textbf{Stage 2.} At this stage, we initialize SyncFormer using the category (1) datasets, where ground-truth depth is provided and camera poses are initialized as identity matrices. We load the Cotracker3~\cite{karaev2024cotracker3} checkpoint to initialize the 2D tracking branch. This stage training takes 3 days of 100k iterations on $8$ H20 GPUs. The video length is shuffle from 12-48 during the training.  
\textbf{Stage 3.} Finally, we fixed the alternation-attention layers in front end and train the whole pipeline in all datasets for 20 hours to converge.

\begin{table*}[t]
    \centering
    \caption{\textbf{3D tracking results on the TAPVid-3D benchmark.} We report the 3D Average Jaccard (AJ), Average 3D Position Accuracy (APD$_{3D}$), and Occlusion Accuracy (OA) across the Aria, DriveTrack, and PStudio subsets. $\textit{offl}^{+}$ and $\textit{offl}^{-}$ denote our offline model with/without considering the camera motion, respectively. COL, Univ2, and Mega are abbreviations for COLMAP, UnidepthV2, and MegaSAM. The \textbf{best} and the \underline{second best} are highlighted.
    }
    \vspace{-1em}
    \label{tab:3d_tracking_tapvid}
    \resizebox{\textwidth}{!}{
    \begin{tabular}{l|l|c|ccc|ccc|ccc|ccc}
        \toprule
        \multirow{2}{*}{\textbf{Methods}} &
        \multirow{2}{*}{\textbf{Type}} &
        \multirow{2}{*}{\makecell{\textbf{Depth /}\\\textbf{Cam Pose}}} &
        \multicolumn{3}{c|}{\textbf{Aria}} &
        \multicolumn{3}{c|}{\textbf{DriveTrack}} &
        \multicolumn{3}{c|}{\textbf{PStudio}} &
        \multicolumn{3}{c}{\textbf{Average}} \\
        & & & AJ $\uparrow$ & APD$_{3D}$ $\uparrow$ & OA $\uparrow$
          & AJ $\uparrow$ & APD$_{3D}$ $\uparrow$ & OA $\uparrow$
          & AJ $\uparrow$ & APD$_{3D}$ $\uparrow$ & OA $\uparrow$
          & AJ $\uparrow$ & APD$_{3D}$ $\uparrow$ & OA $\uparrow$ \\
        \midrule
        BootsTAPIR & Type I & COL & 9.1 & 14.5 & 78.6 & 11.8 & 18.6 & 83.8 & 6.9 & 11.6 & 81.8 & 9.3 & 14.9 & 81.4 \\
        TAPTR & Type I & Univ2 & 15.7 & 24.2 & 87.8 & 12.4 & 19.1 & 84.8 & 7.3 & 13.5 & 84.3 & 11.8 & 18.9 & 85.6 \\
        LocoTrack & Type I & Univ2 & 15.1 & 24.0 & 83.5 & 13.0 & 19.8 & 82.8 & 7.2 & 13.1 & 80.1 & 11.8 & 19.0 & 82.3 \\
        \midrule
        \multirow{3}{*}{CoTracker3} & \multirow{3}{*}{Type I} & COL & 8.0 & 12.3 & 78.6 & 11.7 & 19.1 & 81.7 & 8.1 & 13.5 & 77.2 & 9.3 & 15.0 & 79.1 \\
         &  & Univ2 & 15.8 & 24.4 & 88.9 & 13.5 & 19.9 & 87.1 & 9.2 & 13.8 & 84.2 & 12.8 & 19.4 & 86.7 \\
         &  & Mega & 20.4 & 30.1 &  89.8 & 14.1 & 20.3 & 88.5 & 17.4 & 27.2 & 85.0 & 17.3 & 25.9 & 87.8 \\
        
        \midrule
        \multirow{2}{*}{SpatialTracker} & \multirow{2}{*}{Type II} & Univ2 & 13.6 & 20.9 & 90.5 & 8.3 & 14.5 & 82.8 & 8.0 & 15.0 & 75.8 & 10.0 & 16.8 & 83.0 \\
        & & Mega  & 15.9 & 23.8 & 90.1 & 7.7 & 13.5 & 85.2 & 15.3 & 25.2 & 78.1 & 13.0 & 20.8 & 84.5 \\
        \midrule
        SceneTracker & Type II & Univ2 & - & 23.1 & - & 6.8 & - & - & 12.7 & - & - & - & 14.2 & - \\
        DELTA & Type II & Univ2 & 16.6 & 24.4 & 86.8 & 14.6 & 22.5 & 85.8 & 8.2 & 15.0 & 76.4 & 13.1 & 20.6 & 83.0 \\
        \midrule
        \multirow{2}{*}{Ours-$\textit{offl}^{-}$} & \multirow{2}{*}{Type II} & Univ2 & 18.6 & 26.3 & 90.8 & \underline{16.4} & \underline{24.3} & \underline{90.2} & 18.1 & 27.6 & \underline{86.7} & 17.7 & 26.0 & 89.2 \\
         & & Mega & 22.3 & 32.2 & \underline{93.7} & 15.8 & 23.0 & 90.0 & 18.2 & 28.6 & 87.3 & 18.7 & 27.9 & \underline{}{90.5} \\
        \midrule
        TAPIP3D & Type III & Mega & 23.5 & 32.8 & 91.2 & 14.9 & 21.8 & 82.6 & 18.1 & 27.7 & 85.5 & 18.8 & 27.4 &  86.4 \\
        \midrule
        \multirow{2}{*}{Ours-$\textit{offl}^{+}$} & \multirow{2}{*}{Type III} & Mega & \textbf{24.7} & \textbf{35.2} & \textbf{93.9} & 16.0 & 23.4 & 90.1 & \underline{18.6} & \underline{28.7} & 86.1 & \underline{19.8} & \underline{29.1} & 90.0 \\
         & & Full-ours & \underline{24.6} & \underline{34.7} & 93.6 & \textbf{17.6} & \textbf{26.1} & \textbf{90.8} & \textbf{21.9} & \textbf{32.1} & \textbf{87.4} & \textbf{21.2} & \textbf{31.0} & \textbf{90.6} \\
        \bottomrule
    \end{tabular}
    \vspace{-2em}

    }
\end{table*}

\begin{table*}[t]
    \centering
    \caption{\textbf{Video depth evaluation.} \textbf{Type I} represents the methods specialized in video depth estimation, while \textbf{Type II} are neural reconstruction model, jointly recovering the geometry and camera motion from the video. \textbf{Type III} denotes the SoTA optimization-based method. The \textbf{best} and the \underline{second best} results are highlighted. 
    }
    \vspace{-1em}
    \label{tab:depth_results}
    \resizebox{0.98\textwidth}{!}{%
    \begin{tabular}{l|cc|cc|cc|cc|cc}
        \toprule
        Method / Metrics & \multicolumn{2}{c|}{Average} & \multicolumn{2}{c|}{KITTI~\cite{geiger2013vision}} & \multicolumn{2}{c|}{TUM Dyn \cite{TUM_Dyn}} & \multicolumn{2}{c|}{Bonn \cite{bonn}}  & \multicolumn{2}{c}{Sintel \cite{mayer2016large}} \\
        & AbsRel (↓) & $\delta_{1.25}$ (↑) & AbsRel (↓) & $\delta_{1.25}$ (↑) & AbsRel (↓) & $\delta_{1.25}$ (↑) & AbsRel (↓) & $\delta_{1.25}$ (↑) & AbsRel (↓) & $\delta_{1.25}$ (↑) \\
        \midrule
        DepthCrafter \cite{depthcrafter} & 0.143 & 0.857 & 0.111 & 0.885 & 0.123 & 0.873 & 0.066 & 0.979 & 0.272 & 0.693 \\
        VDA \cite{videoDepth} & 0.154 & 0.882 & 0.080 & 0.951 & 0.118 & 0.920 & 0.049 & \underline{0.982} & 0.370 & 0.674 \\
        \midrule
        DUSt3R~\cite{wang2024dust3r} & 0.240 & 0.766 & 0.124 & 0.849 & 0.187 & 0.792 & 0.174 & 0.835 &  0.475 & 0.591 \\
        MonST3R~\cite{zhang2024monst3r} & 0.171 & 0.802 & 0.083 & 0.934 & 0.197 & 0.726 & 0.061 & 0.954 & 0.343 & 0.594 \\
        CUT3R~\cite{wang2025continuous} & 0.186 & 0.814 & 0.104 & 0.899 & 0.108 & 0.847 & 0.068 & 0.950 & 0.466 & 0.560 \\
        VGGT~\cite{wang2025vggt} & 0.104 & 0.881 & \textbf{0.051} & \underline{0.966} & \underline{0.068} & \underline{0.939} & 0.056 & 0.963 & 0.242 & 0.659 \\
        \midrule
        MegaSAM~\cite{wang2025vggt} & \underline{0.093} & \underline{0.894} & 0.069 & 0.916 & 0.081 & 0.935 & \underline{0.037} & 0.977 & \textbf{0.185} & \textbf{0.746} \\
        \midrule
        Ours & \textbf{0.081} & \textbf{0.910} & \underline{0.052} & \textbf{0.973} & \textbf{0.045} & \textbf{0.976} & \textbf{0.028} & \textbf{0.988} & \underline{0.199} & \underline{0.703} \\
        \bottomrule
    \end{tabular}%
    \vspace{-5em}
    }
\end{table*}

\begin{figure}
\centering

\begin{subfigure}{0.95\linewidth}
    \centering
    \includegraphics[width=\linewidth]{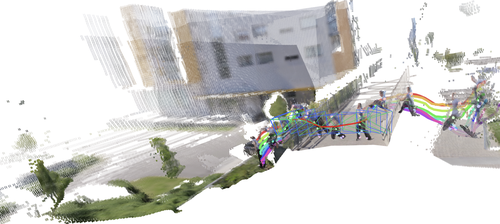}
\end{subfigure}

\vspace{1em}

\begin{subfigure}{0.95\linewidth}
    \centering
    \includegraphics[width=\linewidth]{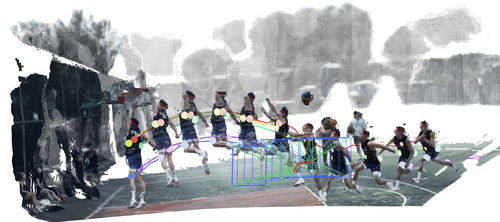}
\end{subfigure}

\caption{
\textbf{Fused Point Clouds, Camera Poses, and 3D Point Trajectories.} 
We visualize the fused point clouds reconstructed from our video depth and camera poses, along with long-term 3D point trajectories in world space.
}
\label{fig:depth_pose}
\end{figure}

\begin{figure}
\centering

\begin{subfigure}{0.95\linewidth}
    \centering
    \includegraphics[width=\linewidth]{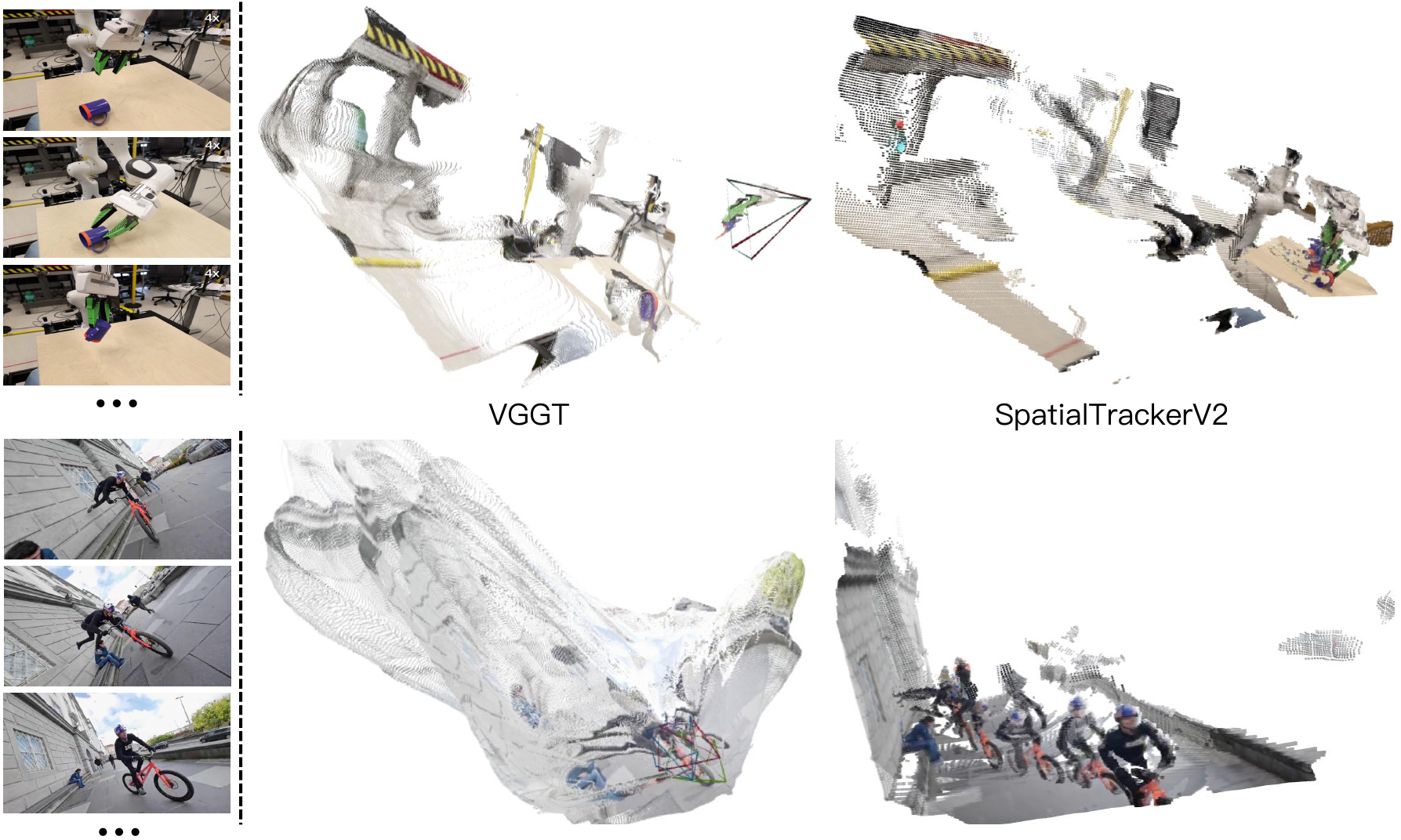}
\end{subfigure}

\caption{
\textbf{Qualitative Comparisons on Internet Videos.} 
To assess generalization, we compare our method with VGGT~\cite{wang2025vggt} on challenging Internet videos.
}
\label{fig:vggt_comp}
\end{figure}

\section{Experiments} 
We evaluate our model across all sub-tasks, including 3D tracking (Sec.~\ref{sec:track3d_exp}), and dynamic 3D reconstruction (Sec.~\ref{sec:dyn_recon}). In addition, we conduct comprehensive ablation studies (Sec.~\ref{sec:ablation}) to analyze the impact of key design choices and to demonstrate the effectiveness of unified modeling and scaling up training.

\subsection{3D Point Tracking}
\label{sec:track3d_exp}
\paragraph{Dataset.} We evaluate our model and compare it with existing baselines on TAPVid-3D~\cite{tap3d}, a comprehensive benchmark spanning diverse scenarios including \textit{Driving}, \textit{Ego-centric}, and \textit{Studio}. This benchmark consists of 4,569 evaluation videos, where the video length varies from 25 to 300 frames, and three 3D point tracking metrics are reported. Specifically, Occlusion Accuracy (OA) measures the precision of occlusion predictions; $\textrm{APD}_{\textit{3D}}$ denotes the average percentage of estimated errors within multiple threshold scales $\delta$; and Average Jaccard (AJ) quantifies the accuracy of both position and occlusion estimation. 

\paragraph{Baselines and Settings.}
\label{sec:exp_settings}
The existing baselines can be broadly categorized into three types.
\textbf{Type I:} 2D trackers followed by depth lifting. We report the current state-of-the-art 2D tracking models, CoTracker3~\cite{karaev2024cotracker3} and BootsTAPIR~\cite{doersch2024bootstap}, as representatives of this category. \textbf{Type II:} 3D trackers in camera space. We include SpatialTrackerV1~\cite{xiao2024spatialtracker} and DELTA~\cite{ngo2024delta}, evaluated with two state-of-the-art depth estimators, \ie, UnidepthV2~\cite{UniDepth} and MegaSAM~\cite{li2025megasam}. Notably, MegaSAM is an optimization-based SLAM system for estimating consistent video depth, which generally performs better than feedforward models such as UnidepthV2. To ensure fair comparisons under the same camera-space 3D tracking setup, our 3D tracker module is also evaluated in combination with these depth estimators.
\textbf{Type III:} 3D trackers in world space. We compare our fully end-to-end model with the recently released TAPIP3D~\cite{zhang2025tapip3d}, which requires consistent video depth and camera poses as input, provided by MegaSAM. For a thorough and fair comparison, we also report results where our depth and pose estimations are replaced by those from MegaSAM.

\paragraph{Quantitative Results and Analysis.}
In the TAPVid-3D benchmark, our method achieves the best results across various settings, consistently outperforming all baselines under comparable conditions as in \cref{tab:3d_tracking_tapvid}. From those detailed comparisons, we can draw several conclusions from it. 
\begin{itemize}
    \item \textbf{Better depth estimation is crucial for 3D tracking:} For Type I methods, the 3D tracking performance largely reflects the accuracy of the underlying 2D tracks and the associated depth predictions. Intriguingly, CoTracker3 + MegaSAM achieves substantial improvements over CoTracker3 + UniDepthV2, with 17.3 vs. 12.8 in AJ, 25.9 vs. 19.4 in $\textit{APD}_{3D}$, and 87.8 vs. 86.7 in OA. This represents the best performance among Type I methods. However, our method Ours-$\textit{offl}^{-}$, which uses the weaker depth estimator UniDepthV2, still outperforms CoTracker3 with MegaSAM, achieving 18.7 vs. 17.3 in AJ and 27.4 vs. 25.9 in $\textit{APD}_{3D}$. This is because Type I methods rely purely on back-projection, making them more sensitive to depth inconsistencies. In contrast, our method directly predicts 3D trajectories using a spatial-temporal transformer, which provides inherent temporal smoothness and robustness to noisy inputs.

    \item 
    \textbf{Camera motion decomposition improves 3D tracking:}  
    As shown in the comparison between Ours-$\textit{offl}^{-}$ + MegaSAM and Ours-$\textit{offl}^{+}$ + MegaSAM, incorporating camera pose estimation clearly improves 3D tracking accuracy. For example, on the Aria subset, our method achieves 24.7 vs. 22.3 in AJ and 35.2 vs. 32.2 in $\textit{APD}_{3D}$. This improvement is largely due to the design of the TAPVid-3D benchmark, which includes a substantial number of background points for evaluation in Aria. These background points primarily reflect camera motion and are particularly challenging for methods that only track in camera space, especially due to frequent out-of-view scenarios. 
    In DriveTrack and Pstudio, the improvements are less, as DriveTrack includes only dynamic points on moving vehicles, while Pstudio contains static scenes with no camera motion. 
    
    \item 
    \textbf{Conclusion:} Our systematic experiments strongly validate our core insight: decomposing 3D point tracking into video depth and camera pose estimation not only enhances each component individually, but also leads to significantly more accurate and robust 3D tracking as a whole, as shown in \cref{fig:depth_pose}. 

\end{itemize}

\subsection{Dynamic 3D Reconstruction}
\label{sec:dyn_recon}
\subsubsection{Video Depth Evaluation}
\label{sec:vid_exp}
\paragraph{Datasets.} To illustrate the effectiveness and generalization ability of our method, we evaluate our video depth estimation method across four mainstream datasets, \ie, KITTI~\cite{geiger2013vision}, Sintel~\cite{mayer2016large}, Bonn~\cite{bonn}, TUM Dynamics~\cite{TUM_Dyn}. 
These datasets encompass both indoor and outdoor scenes, featuring video sequences ranging from $50$ to $110$ frames, providing a comprehensive benchmark for assessing depth estimation consistency across varied environments.

\paragraph{Baselines.} 
We compare our approach against three categories of methods. 
\textbf{Type I} includes SoTA video depth methods, DepthCrafter~\cite{depthcrafter} and Video Depth Anything~\cite{yang2024depth}. 
\textbf{Type II}  consists of SoTA deep reconstruction approaches, \ie DUST3R~\cite{wang2024dust3r}, MonST3R~\cite{zhang2024monst3r}, CUT3R~\cite{wang2025continuous} and VGGT~\cite{wang2025vggt}
\textbf{Type III} denotes the SoTA dynamic Structure-from-Motion (SfM) system, MegaSAM~\cite{li2025megasam}. It leverages an optimization-based paradigm to jointly estimate consistent video depths and camera poses, with constraints enforced by optical flow and monocular depth priors. 

\paragraph{Metrics.} 
We evaluate the performance of video depth estimation using geometric accuracy metrics. To ensure fair comparisons with previous works~\cite{wang2025continuous, zhang2024monst3r, videoDepth}, we follow their approach by aligning the predicted video depth to the ground truth using a shared scale and shift, and then compute the Absolute Relative Error (AbsRel) and $\delta_{1}$ metrics.

\paragraph{Quantitative Results.} As shown in Tab.~\ref{tab:depth_results}, our method clearly outperforms all existing approaches. Specifically, we surpass our baseline in \textbf{Type II}, VGGT~\cite{wang2025vggt}, by a significant margin. On average, our method achieves an AbsRel of 0.081 compared to 0.104 ($\mathbf{-22.1\%}$), and a $\delta_{1.25}$ of 0.910 compared to 0.881 ($\mathbf{+3.3\%}$). Besides, compared to MegaSAM (\textbf{Type III}), our method also maintain clear advantages with 0.081 v.s. 0.093 in AbsRel, and 0.910 v.s. 0.894 in $\delta_{1.25}$. It is important to note that MegaSAM~\cite{li2025megasam} usually needs 5-10 min for a 100 frames video, while our method only takes 5-10 seconds for each which nearly achieves 50 $\times$ faster.

\subsubsection{Camera Poses}
\label{sec:cam_exp}
To evaluate the accuracy of camera motion estimation, we conduct the comparisons on Sintel~\cite{mayer2016large}, TUM Dynamics~\cite{TUM_Dyn}, and Lightspeed~\cite{rockwell2025dynpose}. Sintel and Lightspeed contains numerous challenging dynamic scenes with large ego-motion and object motions, and they are both the synthetic data. TUM Dynamics~\cite{TUM_Dyn} is a real data, captured by well-calibrated RGBD sensors. We report Absolute Translation Error (ATE), Relative Rotation Error (RPE rot) and Relative Translation Error (RTE) after
Sim(3) alignment with the ground truth, as in~\cite{wang2025continuous, zhang2024monst3r}. Shown in \cref{tab:pose_eval_singlecol}, our method outperforms all regression-based methods, much better than our baseline, VGGT~\cite{wang2025vggt} and on par with MegaSAM~\cite{li2025megasam}. Besides, the table illustrates the significance of back end optimization. After joint motion optimization, the pose estimations become nearly twice accurate than before.  

\subsubsection{Internet Videos.}
We present further qualitative comparisons with VGGT~\cite{wang2025vggt} on diverse Internet videos. As illustrated in Fig.~\ref{fig:vggt_comp}, our method achieves more consistent depth and accurate camera poses, showcasing superior generalization.

\begin{table}
    \centering
    \renewcommand{\arraystretch}{1.2}
    \setlength{\tabcolsep}{4pt}
    \caption{
    \textbf{Evaluation on Camera Pose Estimation} on TUM-dynamic~\cite{TUM_Dyn}, Lightspeed~\cite{dai2017scannet}, and Sintel~\cite{mayer2016large}. 
    All values are absolute trajectory error (ATE), relative pose error (RPE) for translation and rotation. The \textbf{best} and the \underline{second best} are highlighted.
    }
    \label{tab:pose_eval_singlecol}
    \resizebox{\linewidth}{!}{%
    \begin{tabular}{l|ccc}
        \toprule
        \textbf{Method} & \textbf{TUM-dynamic} & \textbf{Lightspeed} & \textbf{Sintel} \\
        & (ATE / RPE$_t$ / RPE$_r$) & (ATE / RPE$_t$ / RPE$_r$) & (ATE / RPE$_t$ / RPE$_r$) \\
        \midrule
        Particle-SfM~\cite{Particlesfm}  & -- & 0.185 / 0.075 / 2.990 & 0.129 / 0.031 / 0.535 \\
        Robust-CVD~\cite{RCVD}           & 0.153 / 0.026 / 3.528 & -- / -- / -- & 0.360 / 0.154 / 3.443 \\
        CasualSAM~\cite{CasualSAM}       & 0.071 / 0.010 / 1.712 & -- / -- / -- & 0.141 / 0.035 / 0.615 \\
        DUST3R~\cite{wang2024dust3r}     & 0.140 / 0.106 / 3.286 & 0.412 / 0.177 / 20.100 & 0.290 / 0.132 / 7.869 \\
        CUT3R~\cite{wang2025continuous}  & 0.046 / 0.015 / 0.473 & 0.274 / 0.067 / 1.561 & 0.213 / 0.066 / 0.621 \\
        MonST3R~\cite{zhang2024monst3r}  & 0.098 /  0.019 / 0.935 & 0.149 / 0.046 / \underline{1.210} & 0.078 / 0.038 / 0.490 \\
        VGGT~\cite{wang2025vggt}  & 0.021 /  0.013 / \underline{0.327} & 0.226 / 0.086 / 1.729 & 0.082 / 0.043 / 1.253 \\
        MegaSAM~\cite{li2025megasam}  & \underline{0.013} /  \underline{0.011} / 0.340 & \textbf{0.105} / \underline{0.040} / \textbf{0.996} & \textbf{0.023} / \textbf{0.008} / \textbf{0.060} \\
        \midrule
        \textbf{Ours-Front}                    & 0.038 / 0.022 / 0.480 & 0.203 / 0.079 / 1.689 & 0.075 / 0.045 / 0.805 \\
        \textbf{Ours}                    & \textbf{0.012} / \textbf{0.010} / \textbf{0.305} & \underline{0.134} / \textbf{0.039} / 1.340 & \underline{0.054} / \underline{0.027} / \underline{0.288} \\
        \bottomrule
    \end{tabular}
    }
    \vspace{-1em}
\end{table}

\begin{figure}
    \centering
    \resizebox{\linewidth}{!}{
        \begin{tabular}{c|cccc}
            \includegraphics[width=0.19\linewidth]{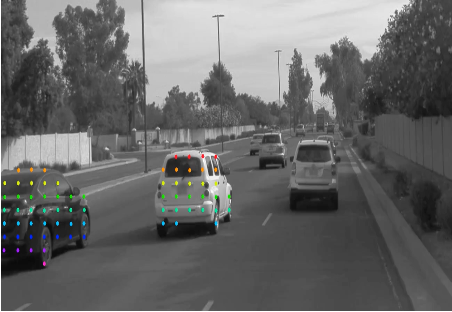} &
            \includegraphics[width=0.19\linewidth]{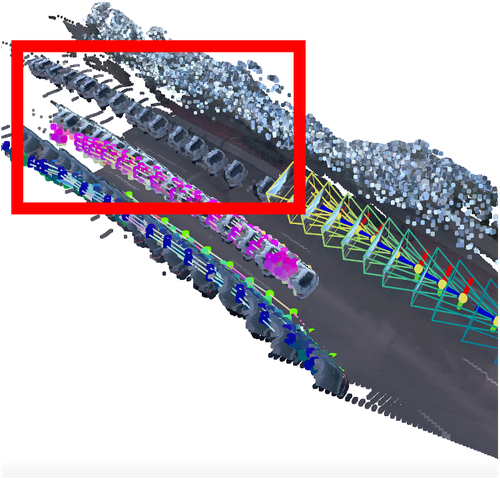} & 
            \includegraphics[width=0.19\linewidth]{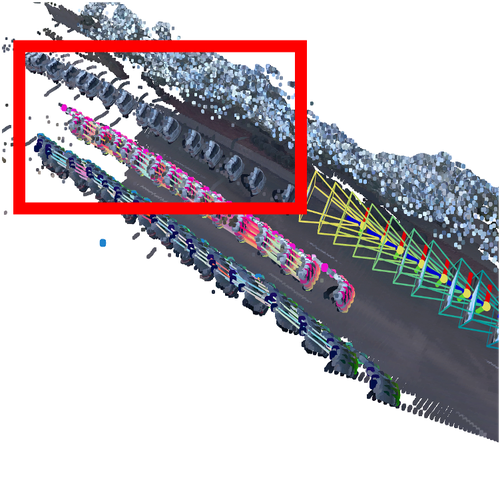} & 
            \includegraphics[width=0.19\linewidth]{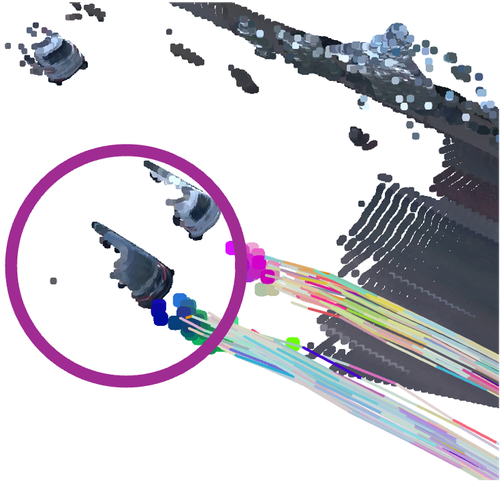} &
            \includegraphics[width=0.19\linewidth]{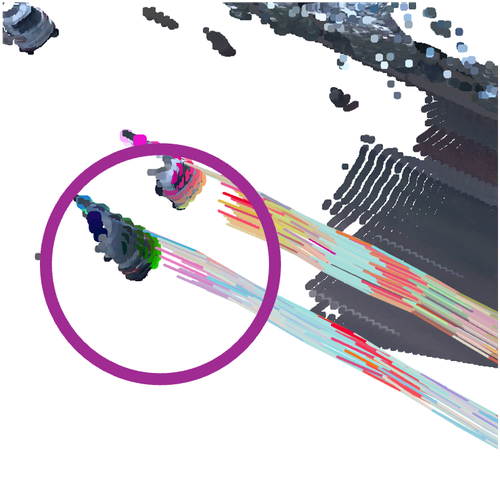} \\
            (a) Query & (b) w/o Joint & (c) w/ Joint & (d) w/o Joint & (e) w/ Joint 
        \end{tabular}
    }
    \caption{The influence of Joint Training.}
    \label{fig:visual_comparison}
\end{figure}

\begin{table}[ht]
\caption{\textbf{TAP-Vid DAVIS results.} Cotracker3-3D is a fine-tuned model by adding simple 3D project layer. Mean represents the average of AJ and $\delta^{\text{vis}}_{\text{avg}}$.}
\label{tab:davis_results}
\resizebox{\linewidth}{!}{%
\begin{tabular}{llcccc}
\toprule
\textbf{Method} & \textbf{Lift 3D} & AJ $\uparrow$ & $\delta^{\text{vis}}_{\text{avg}} \uparrow$ & OA $\uparrow$ & Mean $\uparrow$ \\
\midrule
TAPTR & \ding{55} & 63.0 & 76.1 & 91.1 & 69.5 \\
LocoTrack & \ding{55} & 62.9 & 75.3 & 87.2 & 69.1 \\
CoTracker3 & \ding{55} & 64.4 & 76.9 & \textbf{91.2} & 70.7 \\
\midrule
DELTA & 3D-Dec & 62.7 & 76.7 &  88.2 & 69.7 \\
SpatialTracker & Tri-plane & 61.1 & 76.3 & 89.5 & 68.7 \\
CoTracker3-3D & 3D-Proj & 51.6 & 65.2 & 85.3 & 58.4 \\
Ours & SyncFormer & \textbf{64.9} & \textbf{77.5} & 91.0 & \textbf{71.2} \\
\bottomrule
\end{tabular}
}
\end{table}

\subsection{Ablation Analysis}
\label{sec:ablation}
Our ablation study investigates the impact of training data, joint training, and the SyncFormer design on 2D and 3D point tracking.
For depth estimation, we analyze different loss functions and training strategies to improve the generalizability of video depth prediction.

\paragraph{3D Point Tracking.} To illustrate the gains brought by scaling up the 3D point tracking to a wider range of data, we naively train a base model (Base@K in Table~\ref{tab:rebuttal}), and an advanced one, Base@V,K,P, which is trained on VKITTI, Kubric, and PointOdyssey. As the table shows, Base@V,K,P outperforms Base@K by a clear margin. Meanwhile, our final version is clearly better than these two. Besides, different from Base@K, Base@V,K,P is jointly trained on VKITTI, which brings very significant improvements on a similar type of real data: DriveTrack shows 14.7 vs. 7.4 in AJ and 21.9 vs. 13.3 in APD$_{3D}$. Fig.~\ref{fig:visual_comparison} qualitatively demonstrates the meaning of joint training, i.e., joint training contributes to minimizing the 3D tracking drifts when the model is trained on new patterns of data.

\paragraph{2D Point Tracking.} We report a naive alternative to SyncFormer by modifying the final output layer of CoTracker3~\cite{karaev2024cotracker} to output 2D and 3D tracking. The 2D and 3D embeddings are directly concatenated and projected back to the original dimension using an inserted, learnable linear layer. We name this naive baseline as Cotracker3-3D. As shown in Tab.~\ref{tab:davis_results}, a simple adaptation of 3D lifting leads to a significant drop in 2D tracking accuracy—AJ drops from 64.4 to 51.6, $\delta$ from 76.9 to 65.2, and OA from 91.2 to 85.3. This degradation occurs because the 2D and 3D correlation signals become entangled, disrupting the model's ability to focus on reliable features. Moreover, the update dynamics in 2D and 3D differ substantially, as they occur in separate spaces—UVD space for 2D and camera space for 3D. We also report comparisons with other 3D lifting techniques. These results validate our SyncFormer design, which effectively lifts tracking into 3D while preserving—or even improving—accuracy in 2D.

\paragraph{Depth Estimation.} We ablate different training datasets and their coverage to assess their impact on model performance. Additionally, we study the influence of different loss functions applied to real data, \ie, the pearson loss and full depth losses. As shown in Tab.~\ref{tab:depth_results}, if we applied the full losses for depth will cause the obvious performance drop in the synthetic data, \ie Sintel. On the contrary, if we only used the synthetic data to train the model, we found the performance in the real dataset, especially KITTI will heavily drop. Therefore, we leverage the strategy of using different losses for real and synthetic data to offset the influence of domain gap and different errors distribution. As shown in Tab.~\ref{tab:depth_results}, Ours-Synthetic is our method trained only on the synthetic datasets. Ours-Real-Full is our method using the full depth loss in the real data, where the model was influenced by the noised distributions in the real dataset. The noised of real data has the negative influence on the model's influence on the synthetic data. Meanwhile, the different errors pattern also worsen the model's zero-shot capabilities in the real data. 

\begin{table}[!h]
\centering
\scriptsize
\vspace{-3mm}
\caption{\textbf{Ablation of Training Data and Joint Training.} K, V, P represents Kubric~\cite{greff2022kubric}, VKITTI~\cite{cabon2020vkitti2} and PointOdyssey~\cite{zheng2023pointodyssey} in training. Inference depths are provided by Unidepthv2.} \label{tab:rebuttal}
\vspace{-3mm}
\resizebox{\linewidth}{!}{%
\begin{tabular}{cc|cc|cc|cc}
\toprule
\multirow{2}{*}{{Method}} & \multirow{2}{*}{\makecell{Joint\\Training}} & \multicolumn{2}{c|}{\textbf{Aria}} & \multicolumn{2}{c|}{\textbf{DriveTrack}} & \multicolumn{2}{c}{\textbf{PStudio}} \\
 & &  AJ & APD$_{3D}$ & AJ & APD$_{3D}$ & AJ & APD$_{3D}$ \\
\midrule
Base@{K} & No & 16.0 & 24.4 & 7.4 & 13.3 & 12.6 & 20.3 \\
Base@{V, K, P} & Yes & 15.7 & 24.1 & 14.7 & 21.9 & 17.2 & 27.4 \\
\bottomrule
\end{tabular}
}
\vspace{-2em}
\end{table}

\section{Conclusion}
\label{sec:conclusion}
This work introduces SpatialTrackerV2, a feedforward, scalable, and state-of-the-art approach for 3D point tracking in monocular videos. Built upon a deep exploration of widely-used low- and mid-level representations of motion and scene geometry, our method unifies consistent scene geometry, camera motion, and pixel-wise 3D motion into a fully differentiable end-to-end pipeline. SpatialTrackerV2 accurately reconstructs 3D trajectories from monocular videos, achieving strong quantitative results on public benchmarks and demonstrating robust performance on casually captured Internet videos. We believe SpatialTrackerV2 establishes a solid foundation for real-world motion understanding and brings us a step closer to physical intelligence by exploring large-scale vision data.

\section*{Acknowledgment}
This work was partially supported by Ant Group Research Intern Program, Zhejiang Provincial Natural Science Foundation of China (No. LR25F020003) and Information Technology Center and State Key Lab of CAD\&CG, Zhejiang University.

{
    \small
    \bibliographystyle{ieeenat_fullname}
    \bibliography{main}
}

\end{document}